\documentclass[11pt]{article}  
  
\usepackage{sbc-template}
\usepackage{anyfontsize}
		
\usepackage[vlined,ruled,linesnumbered,portuguese]{algorithm2e}
\usepackage{amssymb}
\usepackage{amsmath}
\usepackage{natbib}
\usepackage{setspace}
\usepackage{graphicx,url}
\usepackage[nolist,nohyperlinks]{acronym}
\usepackage{multirow}
\usepackage[utf8]{inputenc}
\SetKwFor{ParaCada}{para cada}{fazer}{} 
\usepackage{enumitem}

\makeatletter
	\def\hlinewd#1{%
		\noalign{\ifnum0=`}\fi\hrule \@height #1 \futurelet
	  \reserved@a\@xhline}
\makeatother

\title{Métodos de Otimização Combinatória Aplicados ao Problema de Compressão MultiFrases\footnote{Preprint of XLVIII Simpósio Brasileiro de Pesquisa Operacional, 2016.}} 

\begin{acronym}
  \acro{CMF}{Compressão MultiFrases}
  \acro{GP}{Grafo de Palavras}
  \acro{POS}{\textit{Part-Of-Speech}}
  \acro{PLN}{Processamento da Linguagem Natural}
  \acro{SAT}{Sumarização Automática de Textos}
  \acro{LDA}{\textit{Latent Dirichlet Allocation}}
  \acro{TC}{Taxa de Compressão}
\end{acronym}

\author{Elvys Linhares Pontes\inst{1},
Thiago Gouveia da Silva\inst{1,2,3},
Andréa Carneiro Linhares\inst{5},\\
Juan-Manuel Torres-Moreno\inst{1,4},
Stéphane Huet\inst{1}}

\address
 {LIA/CERI -- Université d'Avignon et Pays de Vaucluse (UAPV), Avignon -- France
 \nextinstitute
 Inst. Federal de Educação, Ciência e Tecnologia da Paraíba (IFPB),
 PB -- Brasil
 \nextinstitute
 Instituto de Computação -- Univ. Federal Fluminense (UFF),
 Niterói -- RJ -- Brasil
  \nextinstitute
 Universidade Federal do Ceará (UFC),
 Sobral -- CE -- Brasil
 \email{elvys.linhares-pontes@alumni.univ-avignon.com, thiago.gouveia@ifpb.edu.br }
}

\begin{document}

\maketitle

\newcommand{\XIPT}{\fontsize{11}{15}\selectfont }

\renewenvironment{abstract}{
  \begin{center}
  %\bfseries{ABSTRACT}
  \bfseries ABSTRACT
	\end{center}
    \vspace*{-8pt}
    \hspace*{\parindent}
}

\newenvironment{keywords}{
    %\noindent  \bfseries {KEYWORDS.}
		\noindent  \bfseries KEYWORDS.
}

\newenvironment{resumo1}{
    \begin{center}
    \bfseries RESUMO
    \end{center}
    \vspace*{-8pt}
    \hspace*{\parindent}
}

\newenvironment{palchaves}{
    \noindent\bfseries PALAVRAS CHAVE.
}

\vspace{3mm}
\begin{resumo1}
\fontsize{11}{15}\selectfont{
A Internet possibilitou o aumento considerável da quantidade de informação disponível.
Nesse contexto, a leitura e o entendimento desse fluxo de  informações tornaram-se tarefas dispendiosas.
Ao longo dos últimos anos, com o intuito de ajudar a compreensão dos dados textuais, várias aplicações da área de Processamento de Linguagem Natural (PLN) baseando-se em métodos de Otimização Combinatória vem sendo implementadas.
Contundo, para a \ac{CMF}, técnica que reduz o tamanho de uma frase sem remover as principais informações nela contidas, a inserção de métodos de otimização necessita de um maior estudo a fim de  melhorar a performance da \ac{CMF}.
Este artigo descreve um método de \ac{CMF} utilizando a Otimização Combinatória e a Teoria dos Grafos para gerar frases mais informativas mantendo a gramaticalidade das mesmas.
Um experimento com 40 \textit{clusters} de frases comprova que nosso sistema obteve uma ótima qualidade e foi melhor que o estado da arte.
}
\end{resumo1}

\bigskip
\begin{palchaves}
Otimização Combinatória, Compressão MultiFrases, Grafo de Palavras.

% \bigskip
% \noindent{Área Principal: OA - Outras aplicações em PO }
\end{palchaves}

\vspace{3mm}

\begin{abstract}
\fontsize{11}{15}\selectfont{
The Internet has led to a dramatic increase in the amount of available information. 
In this context, reading and understanding this flow of information have become costly tasks. 
In the last years, to assist people to understand textual data, various Natural Language Processing (NLP) applications based on Combinatorial Optimization have been devised.
However, for Multi-Sentences Compression (MSC), method which reduces the sentence length without removing core information,
the insertion of optimization methods requires further study to improve the performance of MSC.
This article describes a method for MSC using Combinatorial Optimization and Graph Theory to generate more informative sentences while maintaining their grammaticality. 
An experiment led on a corpus of 40 clusters of sentences shows that our system has achieved a very good quality and is better than the state-of-the-art.
}
\end{abstract}

\bigskip

\newpage

\section{Introdução}
\label{sc:int}

  O aumento da quantidade de dispositivos eletrônicos (\textit{smartphones}, \textit{tablets}, etc) e da Internet móvel tornaram o acesso à informação fácil e rápido.
  Através da Internet é possível ter acesso aos acontecimentos de todo o mundo a partir de diferentes \textit{sites}, \textit{blogs} e portais.
  Páginas como a Wikipédia e portais de notícias fornecem informações detalhadas sobre diversas temáticas, entretanto os textos são longos e possuem muitas informações irrelevantes.
  Uma solução para esse problema é a geração de resumos contendo as principais informações do documento e sem redundâncias \citep{elvys:2014}.
  Vista a vasta quantidade e o fácil acesso às informações, é possível automatizar a análise e geração de resumos a partir da análise estatística, morfológica e sintática das frases \citep{torres2014automatic}.

  O \ac{PLN} concerne à aplicação de sistemas e técnicas de informática para analisar a linguagem humana.
  Dentre as diversas aplicações do \ac{PLN} (tradução automática, compressão textual, etc.), a \ac{SAT} consiste em resumir um ou mais textos automaticamente. 
  O sistema sumarizador identifica os dados relevantes e cria um resumo a partir das principais  informações \citep{lia-rag:2015}.
  A \ac{CMF} é um dos métodos utilizados na \ac{SAT} para gerar resumos, que utiliza um conjunto de frases para gerar uma única frase de tamanho reduzido gramaticalmente correta e informativa \citep{filippova:2010,boudin:2013}.

Neste artigo, apresentamos um método baseado na Teoria dos Grafos e na Otimização Combinatória para modelar um documento como um \ac{GP} \citep{filippova:2010} e gerar a \ac{CMF} com uma melhor qualidade informativa.

A seção \ref{sc:fcf} descreve o problema e os trabalhos relacionados à \ac{CMF}.
Detalhamos a abordagem e a modelagem matemática nas seções \ref{sc:nmp} e \ref{sc:mmp}, respectivamente.
O corpus, as ferramentas utilizadas e os resultados obtidos são discutidos na seção \ref{sc:ec}.
Finalmente, as conclusões e os comentários finais são expostos na seção \ref{sc:conc}.

\section{Compressão MultiFrases}
\label{sc:fcf}

A \acf{CMF} consiste em produzir uma frase de tamanho reduzido gramaticalmente correta a partir de um conjunto de frases oriundas de um documento, preservando-se as principais informações desse conjunto.
Uma compressão pode ter diferentes valores de \ac{TC}, entretanto quanto menor a \ac{TC} maior será a redução das informações nele contidas.
Seja $D$ o documento analisado composto pelas frases $\{f_1, f_2, \ldots, f_n\}$ e $frase_{CMF}$ a compressão desse documento, a \ac{TC} é definida por:

\begin{equation}
TC = \frac{||frase_{CMF}||}{\sum_{i = 1}^{n} \frac{||f_i||}{n}}.
\label{eq:tc}
\end{equation}

\noindent onde $||f_i||$ é o tamanho da frase $f_i$ (quantidade de palavras). 
Dessa forma, os principais desafios da \ac{CMF} são a seleção dos conteúdos informativos e a legibilidade da frase produzida. 

Dentre as diversas abordagens feitas sobre a \ac{CMF}, algumas baseiam-se em analisadores sintáticos para a produção de compressões gramaticais.
Por exemplo, \citet{barzilay:2005} desenvolveram uma técnica de geração \textit{text-to-text} em que cada frase do texto é representada como uma árvore de dependência.
De forma geral, essa técnica alinha e combina estas árvores para gerar a fusão das frases analisadas.
Outra abordagem possível é descrita por \citet{filippova:2010}, que gerou compressões de frases de boa qualidade utilizando uma simples modelagem baseada na Teoria dos Grafos e uma lista de \textit{stopwords}\footnote{\textit{Stopwords} são palavras comuns sem relevância informativa para uma frase. Ex: artigos, preposições, etc.}.
\citet{boudin:2013} geraram a \ac{CMF} mais informativas a partir da análise da relevância das frases geradas pelo método de Filippova.

Visto que os trabalhos apresentados utilizaram uma modelagem simples e obtiveram resultados de boa qualidade, este trabalho baseia-se na mesma modelagem utilizada por Filippova e métodos de otimização combinatória para aumentar a informatividade da \ac{CMF}. As subseções \ref{ssc:f} e \ref{ssc:bm} descrevem os métodos utilizados por \citet{filippova:2010} e \citet{boudin:2013}, respectivamente.

\subsection{Filippova}
\label{ssc:f}

\begin{figure}[h]
\centering
\includegraphics[width=15cm]{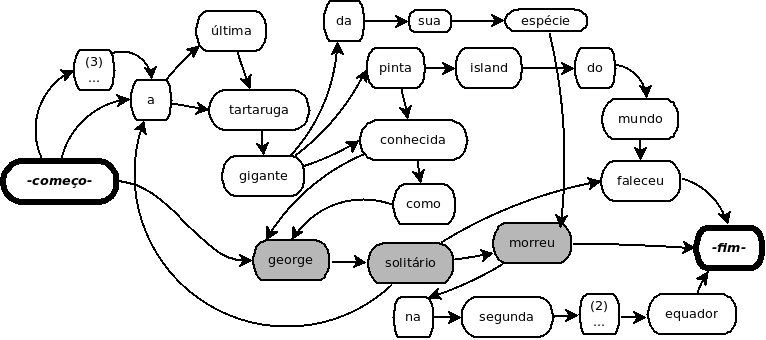}
\caption{Grafo de palavras gerado a partir das frases a-d e um possível caminho representando a compressão \citep{filippova:2010}.
Removemos as vírgulas das frases para facilitar a compreensibilidade do grafo.}
\label{img:gp}
\end{figure}

\citet{filippova:2010} modelou um documento $D$ composto por frases similares como um \acf{GP}. 
O \ac{GP} é um grafo direcionado $GP = (V, A)$, onde $V$ é o conjunto de vértices (palavras) e $A$ é o conjunto de arcos (relação de adjacência). 
Dessa forma, dado um documento $D$ de frases similares $\{f_1, f_2, \ldots, f_n\}$, o \ac{GP} é construído a partir da adição dessas frases no grafo. 
A Figura \ref{img:gp} ilustra o \ac{GP} descrito por Filippova das seguintes frases:
\begin{enumerate}[label=\alph*)]
	\item George Solitário, a última tartaruga gigante Pinta Island do mundo, faleceu.
    \item A tartaruga gigante conhecida como George Solitário morreu na segunda no Parque Nacional de Galapagos, Equador.
    \item Ele tinha apenas cem anos de vida, mas a última tartaruga gigante Pinta conhecida, George Solitário, faleceu.
    \item George Solitário, a última tartaruga gigante da sua espécie, morreu.
\end{enumerate}

Inicialmente, o \ac{GP} é composto pela primeira frase (a) e pelos vértices \textit{-começo-} e \textit{-fim-}. 
Uma palavra é representada por um vértice existente somente se ela possuir a mesma forma minúscula, mesma \ac{POS}\footnote{POS é a classe gramatical de uma palavra numa frase.}, e se não existir outra palavra dessa mesma frase que já tenha sido mapeada nesse vértice. 
Um novo vértice é criado caso não seja encontrado um vértice com suas características no \ac{GP}.
Dessa forma, cada frase representa um caminho simples entre os vértices -começo- e -fim-.

As frases são analisadas e adicionadas individualmente ao \ac{GP}.
Para cada frase analisada, as palavras são inseridas na seguinte ordem:

\begin{enumerate}
	\item Palavras que não sejam \textit{stopwords} e para os quais  não existam nenhum candidato no grafo ou mapeamento não ambíguo;
    \item Palavras que não sejam \textit{stopwords} e para os quais existam vários candidatos possíveis no grafo ou que ocorram mais de uma vez na mesma frase;
    \item \textit{Stopwords}.
\end{enumerate}

Nos grupos 2 e 3, o mapeamento das palavras é ambíguo, pois há mais de uma palavra no grafo que referencia a mesma palavra/POS. 
Nesse caso, as palavras predecessoras e posterioras são analisadas para verificar o contexto da palavra e escolher o mapeamento correto. 
Caso uma dessas palavras não possua o mesmo contexto das existentes no grafo, um novo vértice é criado para representá-la.

Tendo adicionado os vértices, os pesos dos arcos representam o nível de coesão entre as palavras de dois vértices a partir da frequência e da posição dessas palavras nas frases, conforme as Equações \ref{eq:1} e \ref{eq:2}:

\begin{equation}
w(e_{i,j}) = \frac{\textrm{coesão}(e_{i,j})}{freq(i) \times freq(j)},
\label{eq:1}
\end{equation}

\begin{equation}
\textrm{coesão}(e_{i,j}) = \frac{freq(i) + freq(j)}{\sum_{f \in D} dist(f,i,j)^{-1}},
\label{eq:2}
\end{equation}

\begin{equation}
dist(f,i,j) = \left \{
   \begin{array}{l l }
      pos(f,i) - pos(f,j)  & \textrm{se}~ pos(f,i) < pos(f,j) \\
      0   & \textrm{caso contrário}
   \end{array}
   \right .
\label{eq:12}
\end{equation}

\noindent onde $freq(i)$ é a frequência da palavra mapeada no vértice $i$ e a função $pos(f,i)$ retorna a posição da palavra $i$ na frase $f$.

A partir do \ac{GP}, o sistema calcula os 50 menores caminhos\footnote{Ressaltando que cada caminho no GP representa uma frase.} que tenham no mínimo oito palavras e ao menos um verbo.
Por fim, o sistema normaliza os \textit{scores} (distâncias do caminhos) das frases geradas a partir do comprimento das mesmas e seleciona a frase com o menor \textit{score} normalizado como a melhor \ac{CMF}.

\subsection{Boudin e Morin}
\label{ssc:bm}

\citet{boudin:2013} (BM) propuseram um método para melhor avaliar a qualidade de uma frase e gerar compressões mais informativas a partir da abordagem descrita por Filippova (seção \ref{ssc:f}).
BM utilizaram a mesma metodologia de Filippova para gerar os 200 menores caminhos, que tenham no mínimo oito palavras e ao menos um verbo, do \ac{GP}.
Ao invés de realizar uma simples normalização dos valores de cada frase como Filippova, BM mensuraram a relevância da frase gerada (caminho $c$ no GP) a partir das \textit{keyphrases}\footnote{\textit{Keyphrases} são as palavras que representam o conteúdo principal do texto.} e o comprimento das frases, conforme Equações \ref{eq:3} e \ref{eq:4}: 

\begin{equation}
score(c) = \frac{\sum_{i,j \in \textrm{caminho}(c)} w_{(i,j)}}{||c|| \times \sum_{k \in c}score_{kp}(k)},
\label{eq:3}
\end{equation}

\begin{equation}
score_{kp}(k) = \frac{\sum_{w \in k}\textrm{TextRank}(w)}{||k||+1}
\label{eq:4},
\end{equation}

\noindent onde $w_{(i,j)}$ é o score entre os vértices $i$ e $j$ descrito na Equação \ref{eq:1}, o algoritmo TextRank \citep{mihalcea:2004} que calcula a relevância de uma palavra $w$ no \ac{GP} a partir das suas palavras predecessoras e posteriores e $score_{kp}(k)$ é a relevância da \textit{keyphrase} $k$ presente no caminho $c$.
Por fim, a frase com o menor \textit{score} é a escolhida para a compressão do texto.

\section{Nova modelagem do problema}
\label{sc:nmp}

Os métodos de Filippova e BM calculam os menores caminhos do \ac{GP} analisando somente o nível de coesão entre duas palavras vizinhas no texto.
Após a geração dos caminhos, os \textit{scores} de cada caminho são normalizados para escolher o ``menor" deles.
Entretanto, duas palavras possuindo uma forte coesão não significa que as mesmas possuam uma boa informatividade.
Por mais que a normalização ou reanálise das frases seja eficiente, esses métodos estão sempre limitados às frases geradas pela análise do nível de coesão.
Portanto, a geração de 50 ou 200 dos menores caminhos (frases) não garante a existência de uma frase com boa informatividade.
Por isso, propomos um método para analisar concomitantemente a coesão e a relevância das palavras a fim de gerar uma compressão mais informativa de um documento.

O método aqui exposto visa calcular o caminho mais curto analisando a coesão das palavras e bonificando os caminhos que possuam palavras-chaves e \textit{3-grams}\footnote{\textit{3-gram} é formado por 3 palavras vizinhas.} frequentes do texto.
Inicialmente, utiliza-se a mesma abordagem de Filippova (seção \ref{ssc:f}) para modelar um documento $D$ como um \ac{GP} e calcular a coesão das palavras.
Além de considerar a coesão, analisamos as palavras-chaves e os \textit{3-grams} do documento para gerar uma \ac{CMF} mais informativa.
As palavras-chaves auxiliam a geração de caminhos com as principais informações descritas no texto.
Como o documento $D$ é composto por frases similares, consideramos que o documento possui somente uma temática.
A \ac{LDA} é um método para analisar as frases de um texto e identificar o conjunto de palavras que representam as temáticas nele abordadas \citep{lda:2003}.
Configura-se o método \ac{LDA} para identificar o conjunto de palavras que representa uma única temática do documento.
Finalmente, esse conjunto de palavras constitui as palavras-chaves do documento $D$.

Uma outra consideração sobre o documento analisado é que a presença de uma palavra em diferentes frases aumenta sua relevância para a \ac{CMF} (vale salientar que consideramos a relevância dos \textit{stopwords} igual a zero).
% Dessa forma, defini-se a relevância de um \textit{2-gram}\footnote{\textit{2-gram} é formado por 2 palavras vizinhas.} formado pela relevância das palavras dos vértices $i$ e $j$ segundo as Equações \ref{eq:ww} e \ref{eq:wo}.
% \begin{equation}
% \textrm{palavra}(i) = \frac{freq(i)}{||D||} \times \frac{freq_{frases}(i)}{qt_{frases}(D)}
% \label{eq:ww}
% \end{equation}
% \begin{equation}
% \textrm{2-gram}(i,j) = \frac{\textrm{palavra}(i) + \textrm{palavra}(j)}{2},
% \label{eq:wo}
% \end{equation}
% \noindent onde $freq_{frases}(i)$ é quantidade de frases que contém a palavra do vértice $i$, $||D||$ e $qt_{frases}(D)$ são quantidade de palavras e a quantidade de frases do documento $D$, respectivamente.
% A partir da ponderação dos \textit{2-grams}, consideramos que a relevância de um \textit{3-gram} é baseado na relevância dos dois \textit{2-grams} que o formam, como descrito na Equação \ref{eq:tr}:
A partir da ponderação dos \textit{2-grams} (Equação \ref{eq:1}), consideramos que a relevância de um \textit{3-gram} é baseado na relevância dos dois \textit{2-grams} que o formam, como descrito na Equação \ref{eq:tr}:

\begin{equation}
\textrm{3-gram}(i,j,k) = \frac{qt_{3}(i,j,k)}{\max_{a,b,c \in GP} qt_{3}(a,b,c)} \times \frac{w(e_{i,j}) + w(e_{j,k})}{2},
\label{eq:tr}
\end{equation}

\noindent onde $qt_{3}(i,j,k)$ é quantidade de \textit{3-grams} composto pelas palavras dos vértices $i$, $j$ e $k$ no documento.
Os \textit{3-grams} auxiliam a geração de \ac{CMF} com  estruturas importantes para o texto e incrementam a qualidade gramatical das frases geradas.

O nosso sistema calcula os 50 menores caminhos do \ac{GP} que possuam ao menos 8 palavras, baseado na coesão, palavras-chaves e \textit{3-grams} (Equação \ref{fo}). 
Contrariamente ao método de Filippova, as frases podem ter score negativo, pois reduzimos o valor do caminho composto por palavras-chaves e \textit{3-grams}.
Dessa forma, normalizamos os scores dos caminhos (frases) baseado na função exponencial para obter um score maior que zero, conforme a Equação \ref{eq:sn}:

\begin{equation}
score_{norm}(f) = \frac{e^{score_{opt}(f)} }{||f||},
\label{eq:sn}
\end{equation}

\noindent onde $score_{opt}(f)$ é o valor do caminho para gerar a frase $f$ a partir da Equação \ref{fo}. Finalmente, selecionamos a frase com menor score normalizado e contendo, ao menos, um verbo como a melhor compressão das frases do documento.

Para exemplificar nosso método, simplificamos sua análise e utilizamos o texto modelado na Figura \ref{img:gp}.
Nessa figura, existem diversos caminhos possíveis entre os vértices \textit{-começo-} e \textit{-fim-}.
A partir das palavras-chaves ``George", ``gigante", ``solitário"\, ``tartaruga" e ``última", nosso método gerou a compressão ``a tartaruga gigante conhecida george solitário morreu".
Dentre as 5 palavras-chaves analisadas, foi gerada uma compressão contendo 4 delas e com as principais informações das frases.

\section{Modelo Matemático Proposto}
\label{sc:mmp}

Formalmente, o \ac{GP} utilizado pode ser representado como segue: seja $GP=(V,A)$ um grafo orientado simples no qual $V$ é o conjunto de vértices (palavras), $A$ o conjunto de arcos (\textit{2-grams}) e $b_{ij}$ é o peso do arco $(i,j) \in A$ (coesão das palavras dos vértices $i$ e $j$, Equação \ref{eq:1}). 
Sem perda de generalidade, considere $v_0$ como o vértice -começo- e adicione um arco auxiliar do vértice -fim- para $v_0$. 
Adicionalmente, cada vértice possui uma cor indicando se o mesmo é uma palavra-chave. 
Denotamos $K$ como o conjunto de cores em que cada palavra-chave do documento representa uma cor diferente. 
A cor 0 (não palavras-chaves) possui o custo $c_0 = 0$ e as palavras-chaves possuem o mesmo custo $c_k = 1$ (para $k > 0$ e $k \in K$).
O conjunto $T$ é composto pelos \textit{3-grams} do documento com uma frequência maior que 1. 
Cada \textit{3-gram} $t=(a,b,c) \in T$ possui o custo $d_t = \textrm{3-gram}(a,b,c)$ (Equação \ref{eq:tr}) normalizados entre 0 e 1.

Existem vários algoritmos com complexidade polinomial para encontrar o menor caminho em um grafo. 
Contudo, a restrição de que o caminho deve possuir um número mínimo $Pmin$ de vértices (o número mínimo de palavras da compressão) torna o problema NP-Hard. 
De fato, encontrar o menor caminho no GP descrito implica encontrar um ciclo com início e fim em $v_0$, e caso $Pmin$ seja igual a $|V|$, o problema corresponde ao Problema do Caixeiro Viajante (PCV). 
Nesse caso, como o PCV é um caso especial do problema descrito, ele também será NP-Hard.

O modelo matemático proposto para resolução do problema apresentado define cinco grupos de variáveis de decisão:

\begin{itemize}
  \item $x_{ij}$, $\forall (i,j) \in A$, indicando se o arco $(i,j)$ faz parte da solução;
  \item $y_v$, $\forall v \in V$, indicando se o vértice (a palavra) $v$ faz parte da solução;
  \item $z_t$, $\forall t \in T$, indicando se o \textit{3-gram} $t$ faz parte da solução;
  \item $w_k$, $\forall k \in K$, indicando que alguma palavra com a cor (palavra-chave) $k$ foi utilizada na solução; e
  \item $u_v$, $\forall v \in V$, variáveis auxiliares para eliminação de sub-ciclos da solução.
\end{itemize}

O processo de encontrar as 50 melhores soluções se deu pela proibição das soluções encontradas e reexecução do modelo. 
Optamos por essa estratégia em virtude da simetria gerada pela técnica de eliminação de sub-ciclos que utilizamos. 
A formulação é apresentada nas expressões (\ref{fo}) a (\ref{eq13}). 

\begin{align}
  \mathrm{Minimize} ~ 
	\Big( \alpha \sum_{(i,j) \in A} b_{i,j} \cdot x_{i,j}   - \beta \sum_{k \in K} c_k \cdot w_k
  - \gamma \sum_{t \in T} d_k \cdot z_t
    \Big) \label{fo}
\end{align}

\begin{align}
\text{s.a.} \quad \sum_{v \in V} y_v \geq Pmin, \label{eq1} \\
\sum_{v \in V(k)} y_v \geq w_k, && \forall k \in K, \label{eq2} \\
2 z_t \leq x_{ij} + x_{jl}, &&\forall t=(i,j,l)\in T, \label{eq3} \\[7pt]
\sum_{i\in\delta^-(v)} x_{iv} = y_v &&\forall v\in V, \label{eq4} \\
\sum_{i\in\delta^+(v)} x_{vi} = y_v &&\forall v\in V, \label{eq5} \\
y_0 = 1, \label{eq6} %\\[7pt]
\end{align}
\begin{align}
u_0 = 1, \label{eq7} \\[7pt]
u_i - u_j +1 \leq M - M \cdot x_{ij} && \forall (i,j)\in A, j \neq 0, \label{eq8} \\[7pt]
x_{ij} \in \{ 0,1 \}, && \forall (i,j) \in A, \label{eq9} \\[7pt]
z_{l} \in \{ 0,1 \}, && \forall t\in T, \label{eq10} \\[7pt]
y_{v} \in \{ 0,1 \}, && \forall v \in V, \label{eq11} \\[7pt]
w_{k} \in [ 0,1  ], && \forall k \in K, \label{eq12} \\[7pt]
u_{v} \in [1,|V|], && \forall v \in V. \label{eq13}
\end{align}

A função objetiva do programa (\ref{fo}) maximiza a qualidade da compressão gerada. 
As variáveis $\alpha$, $\beta$ e $\gamma$ controlam, respectivamente, a relevância da coesão, das palavras-chaves e dos \textit{3-grams} na geração da compressão.
A restrição (\ref{eq1}) limita o número de vértices (palavras) utilizadas na solução. 
O conjunto de restrições (\ref{eq2}) faz a correspondência entre as variáveis de cores (palavras-chaves) e de vértices (palavras), sendo $V(k)$ o conjunto de todos os vértices com a cor $k$ (uma palavra-chave pode ser representada por mais de um vértice).  
O conjunto de restrições (\ref{eq3}) faz a correspondência entre as variáveis de \textit{3-grams} e de arcos (\textit{2-grams}). 
As igualdades (\ref{eq4}) e (\ref{eq5}) obrigam que para cada palavra usada na solução exista um arco ativo interior (entrando) e um exterior (saindo), respectivamente. 
A igualdade (\ref{eq6}) força que o vértice zero seja usado na solução. 
Por fim, as restrições (\ref{eq7}) e (\ref{eq8}) são responsáveis pela eliminação de sub-ciclos enquanto as expressões (\ref{eq9})-(\ref{eq13}) definem o domínio das variáveis.

Como discutido em \citet{pataki:2003}, existem duas formas clássicas de evitar ciclos em problemas derivados do PCV. 
A primeira consiste na criação de um conjunto exponencial de cortes garantindo que para todo subconjunto de vértices $S \subset V$, $S \neq \emptyset$, haja exatamente $|S|-1$ arcos ativos (mais detalhes em \citet{lenstra:1985}). 
A segunda, conhecida como formulação Miller–Tucker–Zemlin (MTZ) utiliza um conjunto auxiliar de variáveis, uma para cada vértice, de modo a evitar que um vértice seja visitado mais de uma vez no ciclo e um conjunto de arcos-restrições. 
Mais informações sobre a formulação MTZ podem ser obtidas em \citet{oncan:2009}.

Neste trabalho, optamos por eliminar sub-ciclos utilizando o método MTZ, uma vez que sua implementação é mais simples. 
Para tal, utilizamos uma variável auxiliar $u_v$ para cada vértice $v \in V$, e o conjunto de arcos-restrições definido em (\ref{eq8}). 
Nesse grupo de restrições, $M$ representa um número grande o suficiente, podendo ser utilizado o valor $M=|V|$.

\section{Experimentos computacionais}
\label{sc:ec}

O desempenho do sistema proposto foi analisado a partir de diversos valores dos parâmetros ($\beta$ e $\gamma$) associados à função objetivo. 
Os testes foram realizados num computador com processador i5 2.6 GHz e 6 GB de memoria RAM no sistema operacional Ubuntu 14.04 de 64 bits.
Os algoritmos foram implementados utilizando a linguagem de programação Python e as bibliotecas takahe\footnote{Site: http://www.florianboudin.org/publications.html} e gensim\footnote{Site: https://radimrehurek.com/gensim/models/ldamodel.html}. 
O modelo matemático foi implementado na linguagem C++ com a biblioteca Concert e o \textit{solver} utilizado foi o CPLEX 12.6.

\subsection{Corpus e ferramentas utilizadas}

Para avaliar a qualidade dos sistemas, utilizamos o corpus publicado por \citet{boudin:2013}. 
Esse corpus contém 618 frases (média de 33 palavras por frase) divididas em 40 \textit{clusters} de notícias em Francês extraídos do Google News\footnote{Site: https://news.google.fr}.
A taxa de redundância de um corpus é obtida pela divisão da quantidade de palavras únicas pela quantidade de palavras de cada \textit{cluster}. A taxa de redundância do corpus que utilizamos é 38,8\%.
Cada palavra do corpus é acompanhada por sua \ac{POS}.
Para cada \textit{cluster}, há 3 frases comprimidas por profissionais.
Dividimos o corpus em duas partes de 20 \textit{clusters}. 
A primeira parte é utilizada como corpus de aprendizado e a outra parte como corpus de teste.
As frases do corpus de aprendizado tem o tamanho médio de 34,1 palavras e as frases do corpus de teste tem um tamanho médio de 31,6 palavras.

As características mais importante da \ac{CMF} são a informatividade e gramaticalidade das frases. 
A informatividade representa a porcentagem das principais informações transmitidas no texto. 
Como consideramos que as compressões de referência possuem as informações mais importantes, avaliamos a informatividade de uma compressão baseada nas informações em comum entre a mesma e as compressões de referência usando o sistema ROUGE \citep{rouge}.
Utilizamos as métricas de cobertura ROUGE-1 e ROUGE-2, que analisam os \textit{1-grams} e \textit{2-grams}, respectivamente, das compressões de referências presentes nas compressões geradas pelos sistemas, para estimar a informatividade das compressões geradas. 

Devido a complexidade da análise gramatical de uma frase, foi utilizado uma avaliação manual para estimar a qualidade das compressões propostas por nosso sistema.
Como a avaliação humana é lenta, utilizamos essa técnica somente para o corpus de teste.
Para o corpus de aprendizado, decidimos avaliar somente a qualidade informativa (coberturas ROUGE-1 e ROUGE-2) e a \ac{TC} devido ser inviável a análise manual  da quantidade de testes do nosso sistema.

\subsection{Resultados}

Nomeamos nosso sistema como GP+OPT e utilizamos os sistemas de Filippova e de BM como \textit{baselines}.
Testamos o GP+OPT utilizando 1, 3, 5, 7 e 9 palavras-chaves\footnote{Visto que o texto é composto de frases similares sobre um mesmo tópico, consideramos que 9 palavras é a quantidade máxima de palavras-chaves para representar um tópico.} (PC) obtidas a partir do método \ac{LDA}. 
Como o GP+OPT utiliza como base o método de Filippova, tornamos fixo o $\alpha = 1.0$ (priorizando a coesão das compressões geradas) e variamos $\beta$ e $\gamma$ de tal forma que:

\begin{equation}
\beta + \gamma\ <\ 1.0 ;\ \beta,\gamma = 0.0, 0.1, ..., 0.8 , 0.9.
\end{equation}

Todos os sistemas geraram a compressão de um documento em tempo viável (menos de 6 segundos).
Devido à grande quantidade de testes gerados para o corpus de aprendizado, selecionamos os resultados que generalizam o funcionamento do GP+OPT.
A Tabela \ref{tb:cd} descreve a qualidade e a \ac{TC} das compressões.
Essa tabela é dividida em 4 partes. 
A primeira descreve os resultados das \textit{baselines} e as demais partes descrevem os resultados do nosso sistema. 
A primeira parte da tabela comprova que o pós-tratamento utilizado por BM (análise da relevância das \textit{keyphrases}) é melhor que a simples normalização dos scores das frases realizada por  Filippova.
O aumento da relevância dos \textit{3-grams} na nossa modelagem melhora a informatividade da compressão sem aumentar bruscamente a \ac{TC}, pois os \textit{3-grams} favorecem a utilização de \textit{2-grams} frequentes no texto (segunda parte da Tabela \ref{tb:cd}).
Além disso, os 3-grams podem melhorar a qualidade gramatical, pois eles adicionam conjuntos de palavras gramaticalmente corretos à compressão.

Apesar do aumento da relevância das palavras-chaves gerar compressões com uma maior \ac{TC}, as compressões são mais informativas (a terceira parte da tabela) e proporcionam as melhores compressões (linhas em negrito da Tabela \ref{tb:cd}).
Dentre os melhores resultados (última parte da Tabela), escolhemos a versão do nosso sistema com $PC$=9, $\beta$=0.8 e $\gamma$=0.1, pois essa configuração prioriza as palavras-chaves e tenta adicionar \textit{3-grams} às compressões.

\begin{table}[h]
\centering
\caption{A métrica de cobertura do ROUGE e a TC das compressões do corpus de aprendizado.
As linhas em negrito destacam os melhores resultados e a versão selecionada do nosso sistema é marcada por uma estrela.}
\begin{tabular}{lccc}
\hline
Sistemas &	ROUGE-1 &	ROUGE-2 &	TC \\ \hline
\cite{filippova:2010} &	0,58769 &	0,43063 &	51,9\% \\
\cite{boudin:2013} &	0,62364 &	0,45467 &	55,8\% \\ \hline
GP+OPT $PC$=9 $\beta$=0.2 $\gamma$=0.0 &	0,53249 &	0,37515 &	48,3\% \\
% GP+OPT $PC$=9 $\beta$=0.2 $\gamma$=0.1 &	0,58589 &	0,42311 &	49,26\% \\
GP+OPT $PC$=9 $\beta$=0.2 $\gamma$=0.2 &	0,55202 &	0,40276 &	50,0\% \\
% GP+OPT $PC$=9 $\beta$=0.2 $\gamma$=0.3 &	0,61067 &	0,44680 &	50,87\% \\
GP+OPT $PC$=9 $\beta$=0.2 $\gamma$=0.4 &	0,57806 &	0,42385 &	51,8\% \\
% GP+OPT $PC$=9 $\beta$=0.2 $\gamma$=0.5 &	0,61630 &	0,45347 &	51,76\% \\
GP+OPT $PC$=9 $\beta$=0.2 $\gamma$=0.6 &	0,58996 &	0,43265 &	54,1\% \\ \hline
% GP+OPT $PC$=9 $\beta$=0.2 $\gamma$=0.7 &	0,62251 &	0,45912 &	54,98\% \\ \hline
GP+OPT $PC$=9 $\beta$=0.0 $\gamma$=0.2 &	0,49858 &	0,36156 &	43,0\% \\
% GP+OPT $PC$=9 $\beta$=0.1 $\gamma$=0.2 &	0,56524 &	0,41525 &	46,18\% \\
GP+OPT $PC$=9 $\beta$=0.2 $\gamma$=0.2 &	0,55202 &	0,40276 &	50,0\% \\
% GP+OPT $PC$=9 $\beta$=0.3 $\gamma$=0.2 &	0,61472 &	0,45273 &	51,61\% \\
GP+OPT $PC$=9 $\beta$=0.4 $\gamma$=0.2 &	0,58039 &	0,42394 &	52,1\% \\
% GP+OPT $PC$=9 $\beta$=0.5 $\gamma$=0.2 &	0,61892 &	0,45429 &	54,54\% \\
GP+OPT $PC$=9 $\beta$=0.6 $\gamma$=0.2 &	0,60072 &	0,43884 &	54,3\% \\ \hline
GP+OPT $PC$=9 $\beta$=0.2 $\gamma$=0.7 &	0,59956 &	0,44160 &	48,3\% \\
GP+OPT $PC$=7 $\beta$=0.6 $\gamma$=0.3 &	0,59981 &	0,43467 &	48,3\% \\
GP+OPT $PC$=9 $\beta$=0.6 $\gamma$=0.3 &	0,61707 &	0,45033 &	48,3\% \\
\textbf{GP+OPT $PC$=9 $\beta$=0.8 $\gamma$=0.0} &	\textbf{0,62874} &	\textbf{0,46089} &	56,6\% \\
\textbf{GP+OPT $PC$=9 $\beta$=0.8 $\gamma$=0.1}$^\star$ &	\textbf{0,62874}$^\star$ &	\textbf{0,46089}$^\star$ &	56,6\% \\
\textbf{GP+OPT $PC$=9 $\beta$=0.9 $\gamma$=0.0} &	\textbf{0,62874} &	\textbf{0,46089} &	56,6\% \\ \hline
\end{tabular}
\label{tb:cd}
\end{table}

Selecionado a melhor configuração do nosso sistema, validamos a qualidade dos sistemas utilizando o corpus de teste (Tabela \ref{tb:ct}).
Similar aos resultados do corpus de aprendizado, o método de BM foi melhor que o método de Filippova para as métricas ROUGE-1 e ROUGE-2. 
GP+OPT obteve resultados bem superiores que as \textit{baselines} comprovando que a análise da coesão juntamente com as palavras-chaves e \textit{3-grams} auxiliam a geração de melhores compressões.
Apesar dos valores da \ac{TC} do nosso sistema terem sido maiores que os valores da \ac{TC} das \textit{baselines}\footnote{A diferença do tamanho médio das frases entre os sistemas GP+OPT e Filippova foi 3,7 palavras.}, a \ac{TC} do sistema GP+OPT ficou próxima da $TC$ das compressões dos profissionais (\ac{TC} = 59\%).

\begin{table}[h]
\centering
\caption{A métrica de cobertura do ROUGE e a TC das compressões do corpus de teste.}
\begin{tabular}{lccc}
\hline
Sistemas &	ROUGE-1 &	ROUGE-2 &	TC \\ \hline
\cite{filippova:2010} &	0,58455 &	0,43939 &	51,1\% \\
\cite{boudin:2013} &	0,62116 &	0,45734 &	55,2\% \\
\textbf{GP+OPT $PC$=9 $\beta$=0.8 $\gamma$=0.1} &	\textbf{0,70009} &	\textbf{0,50207} &	65,1\% \\
\hline
\end{tabular}
\label{tb:ct}
\end{table}

Com o intuito de melhor analisar a qualidade informativa e gramatical das compressões, 5 franceses avaliaram as compressões geradas por cada sistema e notificaram a qualidade gramatical e informativa para o corpus de teste (Tabela \ref{tb:am}).
Nosso sistema gerou estatisticamente compressões mais informativas que as \textit{baselines}.
Apesar da média da gramaticalidade do nosso sistema ter sido inferior a dos demais sistemas, não podemos confirmar qual sistema é estatisticamente melhor para a gramaticalidade devido ao fato dos intervalos de confiança da gramaticalidade dos sistemas se cruzarem.
Portanto, nosso sistema pode gerar compressões com qualidade gramatical igual às compressões geradas pelos métodos de Filippova ou de BM.

Desse modo, pode-se afirmar que o GP+OPT apresentou melhores resultados que as \textit{baselines} gerando compressões mais informativas e com uma boa qualidade gramatical.

\begin{table}
\centering
\caption{Média e intervalo de confiança da avaliação manual da informatividade e gramaticalidade das compressões do corpus de teste.
As notas possíveis para cada métrica são entre 0 e 5.
}
\begin{tabular}{lcc}
\hline
Sistemas &	Gramaticalidade & Informatividade \\ \hline
\cite{filippova:2010} &	\textbf{4,2} $\pm$ 0,18 &	2,86 $\pm$ 0,32\\
\cite{boudin:2013} & 3,99 $\pm$ 0,21 &	3,31 $\pm$ 0,32\\
\textbf{GP+OPT $PC$=9 $\beta$=0.8 $\gamma$=0.1} & 3,93 $\pm$ 0,22 & \textbf{3,95} $\pm$ 0,23 \\
\hline
\end{tabular}
\label{tb:am}
\end{table}

\section{Considerações Finais e Proposta de Trabalhos Futuros}
\label{sc:conc}

A \ac{CMF} gera frases de boa qualidade sendo uma ferramenta interessante para a SAT.
A análise concomitante da coesão, palavras-chaves e \textit{3-grams} identificaram as informações principais do documento.
Apesar do nosso sistema ter gerado compressões com uma \ac{TC} maior que as \textit{baselines}, a informatividade foi consideravelmente melhor.
A análise manual dos franceses comprovou que nosso método gerou compressões mais informativas e mantendo uma boa qualidade gramatical.

Os próximos trabalhos visam criar um corpus similar ao de BM para o idioma Português e testar o desempenho do nosso sistema para diferentes idiomas.
Além disso, pretende-se adaptar o sistema para escolher a relevância das palavras-chaves e dos \textit{3-grams} baseados no tamanho e no vocabulário do documento.
Finalmente, objetiva-se implementar diferentes métodos para a obtenção de palavras-chaves, a fim de avaliar o impacto de cada um na qualidade da geração da \ac{CMF}.

\section*{Agradecimentos}

Este trabalho foi financiado parcialmente pelo projeto europeu CHISTERA-AMIS ANR-15-CHR2-0001.

\newpage

\bibliographystyle{sbpo}
\bibliography{references}

\end{document}